\documentclass[letterpaper, 10 pt, conference]{ieeeconf}                   
\IEEEoverridecommandlockouts
\usepackage{url}

\usepackage{savesym}
\usepackage{amsmath,amssymb,amsfonts}
\usepackage{algorithm}
\usepackage[noend]{algpseudocode}
\usepackage{comment}
\usepackage{xspace}
\usepackage{scalerel}
\usepackage{wrapfig}
\usepackage[font=footnotesize]{subcaption}
\usepackage[font=scriptsize]{caption}
\usepackage[left]{showlabels}
\usepackage{multicol}
\usepackage[bookmarks=true]{hyperref}
\let\labelindent\relax
\usepackage{enumitem}
\usepackage[table,xcdraw]{xcolor}
\usepackage{tabularx}

\usepackage{enumitem,amssymb}
\newlist{todolist}{itemize}{2}
\setlist[todolist]{label=$\square$}

\usepackage[textsize=scriptsize,textwidth=3.cm]{todonotes}

\graphicspath{
	{./figs/}
}

\hypersetup{letterpaper,bookmarksopen,bookmarksnumbered,
pdfpagemode=UseOutlines,
colorlinks=true,
linkcolor=blue,
anchorcolor=blue,
citecolor=blue,
filecolor=blue,
menucolor=blue,
urlcolor=blue
}

\newcommand{\ignore}[1]{}

  \def\H{\mathcal{H}}
  \def\L{\mathcal{L}}

  \def\L{\mathcal{L}}
  
\def\D{\mathcal{D}} \def\W{\mathcal{W}} 
\def\M{\mathcal{M}}  
 \def\L{\mathcal{L}}

\newcommand{\Cpp}{C\raise.08ex\hbox{\tt ++}\xspace}

\makeatletter
\def\thmhead@plain#1#2#3{%
  \thmname{#1}\thmnumber{\@ifnotempty{#1}{ }\@upn{#2}}%
  \thmnote{ {\the\thm@notefont#3}}}
\let\thmhead\thmhead@plain
\makeatother

\newcommand\plr{\textup{PLR}\xspace}

\newcommand\NN{\textup{NN}}

\newcount\Comments  
\definecolor{darkgreen}{rgb}{0,0.5,0}
\definecolor{darkred}{rgb}{0.7,0,0}
\definecolor{teal}{rgb}{0.3,0.8,0.8}
\definecolor{orange}{rgb}{1.0,0.5,0.0}
\definecolor{purple}{rgb}{0.8,0.0,0.8}
\newcommand{\kibitz}[2]{\ifnum\Comments=1{\textcolor{#1}{\textsf{\footnotesize #2}}}\fi}

\begin{document}

\title{Planning, Learning and Reasoning Framework for Robot Truck Unloading}

\author{ Fahad Islam$^{*1}$,
  Anirudh Vemula$^{*1}$,
  Sung-Kyun Kim$^{2}$,
  Andrew Dornbush$^{1}$,
  Oren Salzman$^{3}$,
    Maxim Likhachev$^{1}$
    \thanks{
      $^*$ Equal contribution. Alphabetical ordering}
    \thanks{
	    $^{1}$Carnegie Mellon University, 
	    5000 Forbes Avenue, Pittsburgh, PA, USA}
    \thanks{
	    $^{2}$NASA Jet Propulsion Laboratory, California Institute of Technology,
	    4800 Oak Grove Drive, Pasadena, CA, USA}
    \thanks{
        $^{3}$Technion-Israel Institute of Technology,
	    Haifa, 3200003, Israel
	    }
    }

    \maketitle

\begin{abstract}
We consider the task of autonomously unloading boxes from trucks using
an industrial manipulator robot. There are multiple challenges that
arise: (1)~real-time motion planning for a complex robotic system
carrying two articulated mechanisms, an arm and a scooper, (2)
decision-making in terms of what action to execute next given
imperfect information about boxes such as their masses, (3)
accounting for the sequential nature of the problem where current
actions affect future state of the boxes, and (4)~real-time
execution that interleaves high-level decision-making with lower level
motion planning.  In this work, we propose a planning, learning, and
reasoning framework to tackle these challenges, and describe its
components including motion planning, belief space
planning for offline learning, online decision-making based on offline
learning, and an execution module to combine decision-making with motion
planning. We analyze the performance of the framework on real-world
scenarios. In particular, motion planning and execution modules are
evaluated in simulation and on a real robot, while offline learning
and online decision-making are evaluated in simulated real-world scenarios. Video of our physical robot experiments can be found at \url{https://www.youtube.com/watch?v=hRiRhS0kgSg}
\end{abstract}

\section{Introduction}
\label{sec:intro}
\begin{figure}[t]
\centering
\begin{subfigure}{.65\columnwidth}
  \includegraphics[clip,trim=0cm 0cm 5cm 0cm, width=\linewidth]{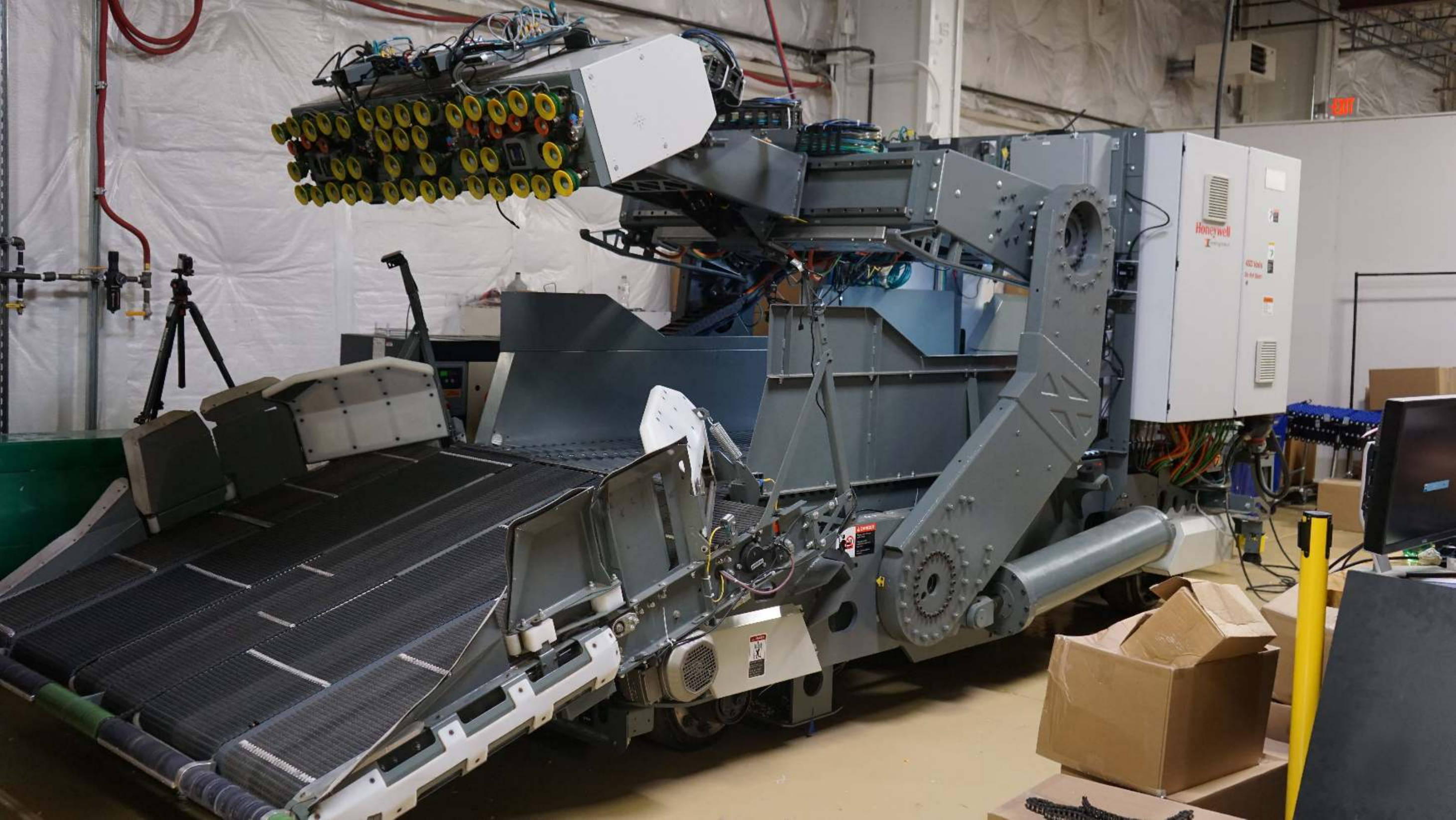}
  \caption{}
  \label{fig:robot}
\end{subfigure}
\hspace{3mm}
\begin{subfigure}{.25\columnwidth}
  \centering
  \includegraphics[width=\linewidth]{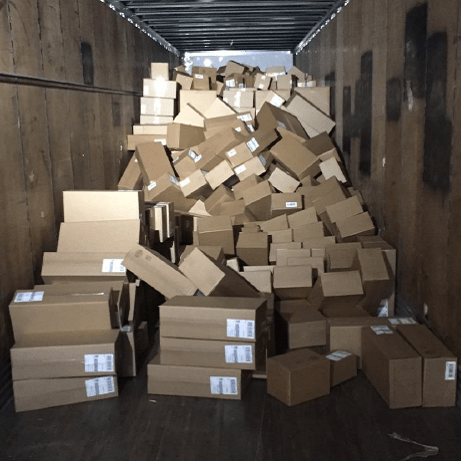}\\
  \includegraphics[width=\linewidth]{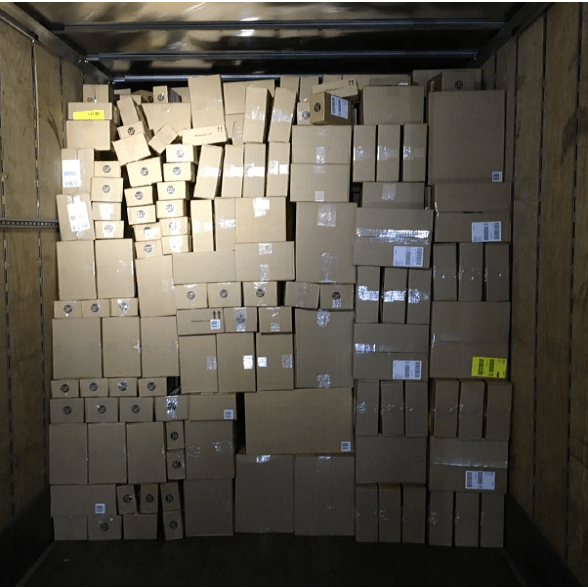}
  \caption{}
  \label{fig:trailers}
\end{subfigure}
\caption{
\protect (\subref{fig:robot}) Truck-unloader robot with its
manipulator-like (arm) and scooper-like (nose) end effectors.
\protect (\subref{fig:trailers}) Examples of different trailers that need to be unloaded.}
\label{fig:gmt}
\vspace{-7mm}
\end{figure}

Industrial automation has improved efficiency and decreased costs for
modern industries, owing largely to the introduction of 
robots into factories \cite{Graetz2015RobotsAW}. In this work, we
consider warehouse automation 
and tackle the problem of automated truck
unloading. Unloading boxes from trucks is a daily operation in
warehouses that requires manual labor
and could benefit from automation \cite{Black2019}, which can help increase 
throughput, i.e., rate of unloading boxes, and reduce employee work-related injuries.
To achieve this goal, we propose a framework to plan robust actions for a
custom-built truck-unloading robot (Fig.~\ref{fig:robot}) equipped
with a mobile omnidirectional base referred to as \textit{base} and two articulated mechanisms--a manipulator-like tool with suction
grippers and a  scooper-like tool with
conveyor belts, referred to as \textit{arm} and \textit{nose}, respectively. 
The objective of the
task is to unload boxes as quickly and efficiently as possible,
without damaging the boxes or the robot in realistic truck
environments (Fig.~\ref{fig:trailers}.)

Achieving high-level goals such as unloading a trailer involves both
high-level decision making and low-level motion planning. Typically,
such problems
are formulated in deterministic settings where states
are fully observable and outcomes of actions are deterministic. Task
planning is then performed on logical predicates which require
geometric reasoning to be applicable to
real-world problems~\cite{SFRCRA14}. Due to the complexity that comes
from hierarchical planning, these approaches often suffer from long
computational times and thus are not well suited for time-critical
applications such as truck unloading. Kaelbling and
Lozano-P{\'e}rez~\cite{kaelbling2013integrated} accounted for
uncertainty when planning in non-deterministic settings but this comes at a
cost of additional computational overhead.  
The computational complexity can be reduced, to some degree, by
learning to predict solution constraints on the
planner~\cite{kim2017learning}. This approach allows to prune the
search space but does not reason about the high-level sequential
decision making for multi-step tasks. In contrast to these approaches,
we use offline planners to precompute offline strategies and then use
online decision making techniques to choose appropriate strategies in real time.
This approach is similar to using slow offline
planners to generate data to train online policies~\cite{guo2014deep,
silver2016mastering}.

Our domain is non-deterministic as we do not have perfect
information about the world (e.g., box masses).
Thus, we require planners capable of
handling uncertainty. Planning under observation and model uncertainty
is usually framed as a Partially Observable Markov Decision Process
(POMDP)~\cite{KLC98,K15}. Many POMDP solvers use point-based value
iteration that requires explicit models for the probability
distributions~\cite{pineau2006anytime,ross2008online}.  
Another approach to solving POMDPs is by running  Monte-Carlo searches
on a simulator~\cite{SV10} that  does not rely on explicit
probability distributions. In our work, we combine this approach with advanced
heuristic-search methods \cite{KSL19} to efficiently plan robust actions.

There are several challenges in automated truck unloading: 
(1)~requiring real-time motion planning capabilities to ensure high throughput, 
(2)~dealing with imperfect information about the world, such as box masses, 
(3)~accounting for sequential nature of the task where current actions affect future box configurations, and finally 
(4)~ensuring real-time execution on the robot interleaving high-level
decision making and low-level motion planning. The proposed framework
is designed to tackle these challenges.

We employ a high-fidelity simulator V-REP \cite{VREP2013}, that is capable of simulating the
robot, the trailer and boxes. This simulator is used by the
\textit{Strategy Generator} module in the
offline phase to compute high-level plans,
\textit{strategies}, for sampled environments accounting for
uncertainty regarding the box masses while maximizing
throughput. However, the space of environments is too large to use
these precomputed strategies online without any adaptation. To this
end, we learn a mapping, which we call the \textit{Strategy Chooser}, that chooses the most relevant
strategy among precomputed strategies given current boxes
configuration. This enables our
framework to generalize to previously unseen environments online.
During execution, the \textit{Motion Planner}
module receives queries, and uses a state-of-the-art
motion planner to plan joint trajectories for the robot minimizing
execution time.
Finally, the task of instantiating high-level planned strategies into low-level
motion planning queries to move the robot's end-effectors is done by the \textit{Strategy Executor} 
module which ensures real-time execution. The framework is briefly
summarized in Fig.~\ref{fig:sys}. While the framework described above
is general and we anticipate that 
it can be applied to many domains, we focus on
its performance for our specific
application of automated truck unloading.
\vspace{-1mm}
\section{PLR Framework}
\label{sec:architecture}
\vspace{-1mm}
The planning, learning and reasoning (PLR) framework is tasked with
planning collision-free motions of the robot that will maximize the
throughput i.e., rate of
boxes unloaded, while avoiding damage to boxes and the robot.
It continuously receives as an input, the current perceived world state~$\W$
 of the environment provided by the robot's perception system, and outputs a trajectory to be
executed by the robot's actuators.
The world state~$\W$, contains 
(1)~a voxel grid estimating the volume occupied by the trailer's walls, ceiling and floor,
and
(2)~an estimate of the position and orientation of all the
perceived boxes in the trailer.
Box masses are not
provided, and they are the main sources of uncertainty in~$\W$. 
We will also use the notion of the true world state~$\W^t$, 
which contains information about all the boxes in the
trailer including their masses.

The robot is equipped with a wide suite of sensors, such as RGB and depth
sensors, that allow it to
estimate occupancy of the trailer's walls, and estimate box poses
which are used to construct~$\W$. In addition to the
real robot, we also employ a simulation of the
robot, trailer, and boxes to estimate the outcomes of actions, and
perform offline learning. The sensors are not simulated and 
we assume ground truth box poses and trailer occupancy for simulation
results. A discussion of robot's sensor suite and perception module is outside the scope of
this paper.

Before we detail the framework's modules, we introduce two key notions
that drive our framework---\emph{strategies}, and \emph{abstract actions}.

\begin{figure}[t]
\centering
   \hspace*{-2.0em}
  \begin{subfigure}{.63\columnwidth}
	\centering
    \includegraphics[height=3.cm]{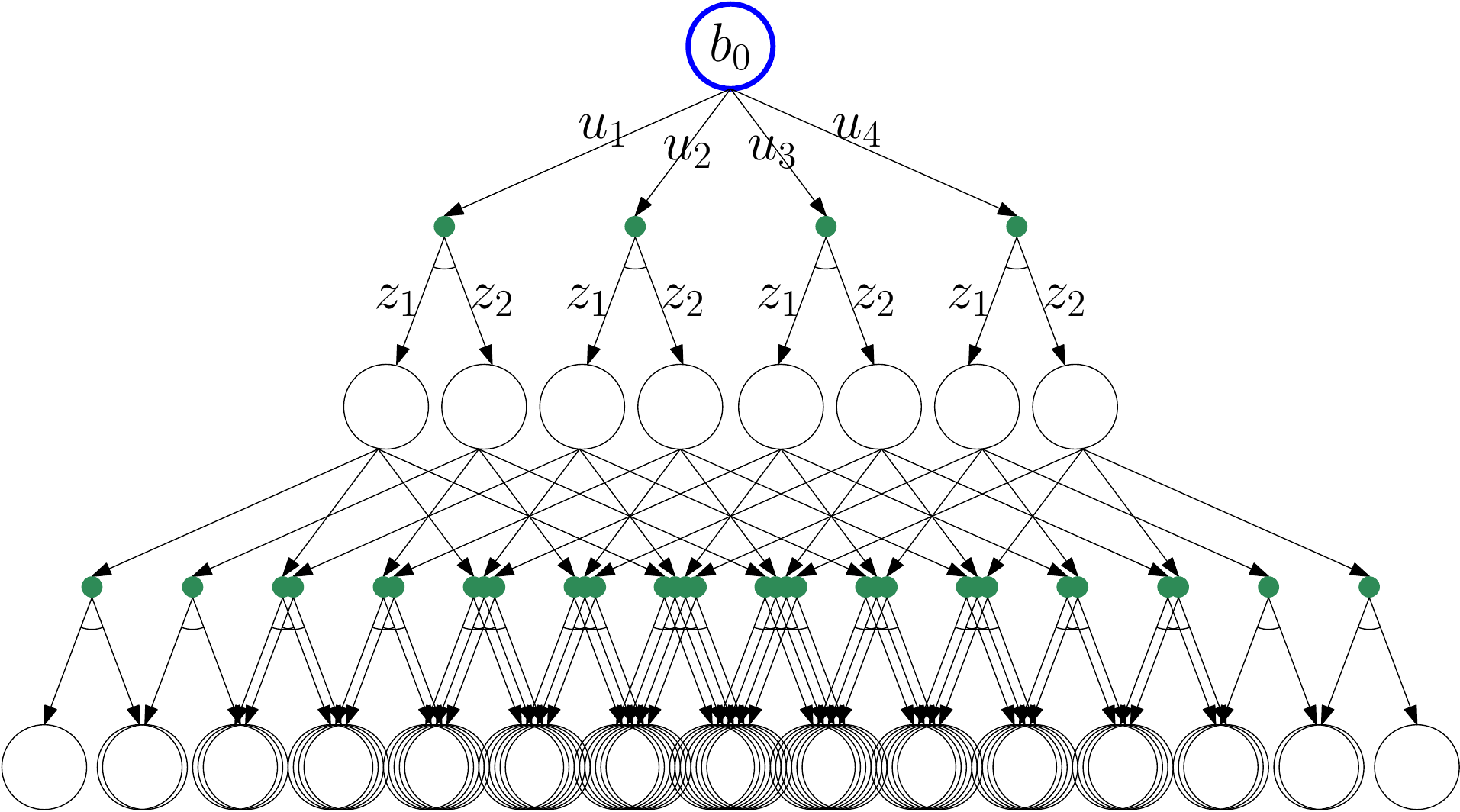}
    \caption{}
    \label{fig:tree}
  \end{subfigure}
  \hspace*{-1.2em}
  \begin{subfigure}{.33\columnwidth}
	\centering
    \includegraphics[height=3.cm]{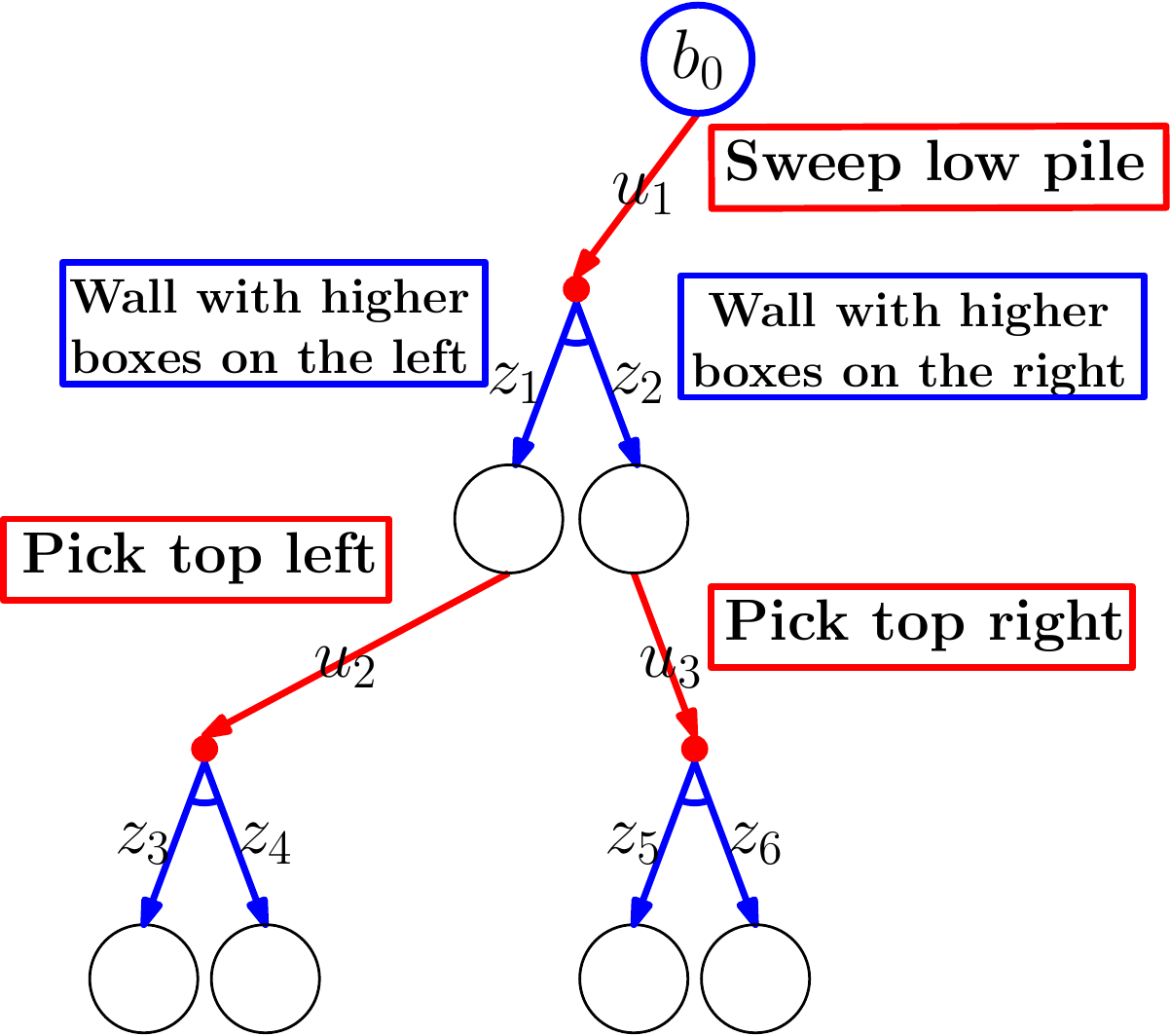}
    \caption{}
    \label{fig:treesol}
  \end{subfigure}
	\caption{
	\protect (\subref{fig:tree}) Depiction of a belief tree constructed over actions ($u_1$ through $ u_4$) and observations ($z_1$ and $z_2$) from an initial belief $b_0$.
  \protect (\subref{fig:treesol}) Example of a decision tree, or \textit{strategy}, that specifies the best actions (red arrows) along the possible action/observation histories from the initial belief $b_0$.}
	\label{fig:trees}
\vspace{-2mm}
\end{figure}

\subsection{Strategies and Abstract Actions} 
\label{sec:key-ingredients}
We represent a robust plan as a decision tree of sequential
actions and observations which is referred to as a \textit{strategy}
hereinafter (Fig.~\ref{fig:treesol}). 
Due to the imperfect control and limited sensing capabilities of the
robot, and the stochastic nature of environment dynamics,
execution of the same action sequence may result in different
outcomes.
Thus, a strategy should specify which action to take on each possible outcome along the sequence, and
the decision tree is an adequate data structure to encode such information.

More specifically, a strategy is a mapping that describes
which action to take given a world state $\W$.
Since the robot does not have access to the true state $\W^t$, it has to infer the state from 
the \textit{initial belief} (probabilistic estimation of the initial state) and 
the \textit{history} of the executed actions and obtained observations.
Given a strategy that contains the initial belief and the history of previous actions and corresponding observations, the robot
can effectively infer the current true state and execute the best
action.

\begin{figure}[t]
\centering
  \begin{subfigure}{.46\columnwidth}
	\centering
    \includegraphics[clip,trim=0cm 1cm 0cm 1cm,height=2.5cm]{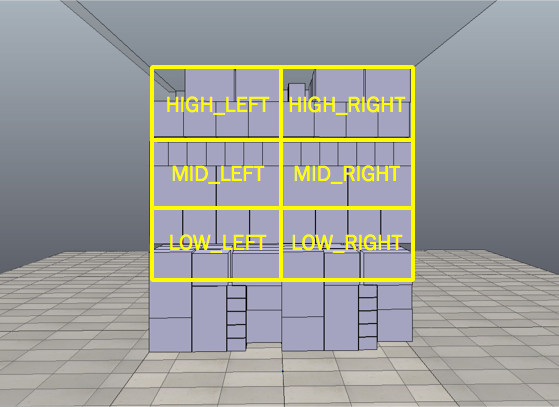}
    \caption{Pick action parameters.}
    \label{fig:abspickact}
  \end{subfigure}
  \begin{subfigure}{.52\columnwidth}
	\centering
    \includegraphics[height=2.5cm]{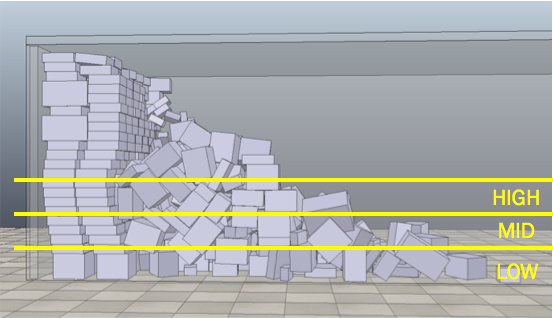}
    \caption{Sweep action parameters.}
    \label{fig:abssweepact}
  \end{subfigure}
	\caption{
    Illustration of abstract action parameters for picking and sweeping.
    An abstract Pick action is specified by the \texttt{height} and \texttt{side} parameters, and an abstract Sweep action is specified by the \texttt{height} parameter.
    These abstract actions are instantiated based on the latest observation at runtime, which makes the action execution to be adaptive to the percieved world state.
  }
	\label{fig:absact}
\vspace{-7mm}
\end{figure}

We present another important notion in our framework, \textit{abstract action}.
Note that a strategy is described by actions and observations at a semantic level.
Since the task is highly complex, we adopt a hierarchical motion
planning scheme to reduce the complexity of the problem, and an
abstract action serves as a macro in the architecture.

Abstract actions are primitive actions at a semantic level, but they
are instantiated at runtime according to the latest world state.
In this work, we have two types of abstract actions, Pick and Sweep, and each abstract action has additional parameters for instantiation (Fig.~\ref{fig:absact}).
For example, Pick with \texttt{height} and \texttt{side}
parameter values of \texttt{high} and \texttt{left} will be
instantiated with real values at runtime as a sequence of actions to
pick the boxes from the high, left part of the wall. This enables our
framework to learn offline strategies that are applicable for a wide
variety of environments.

\begin{figure*}[t]
\centering
  \begin{subfigure}{.31\textwidth}
    \centering
    \includegraphics[width=0.9\linewidth]{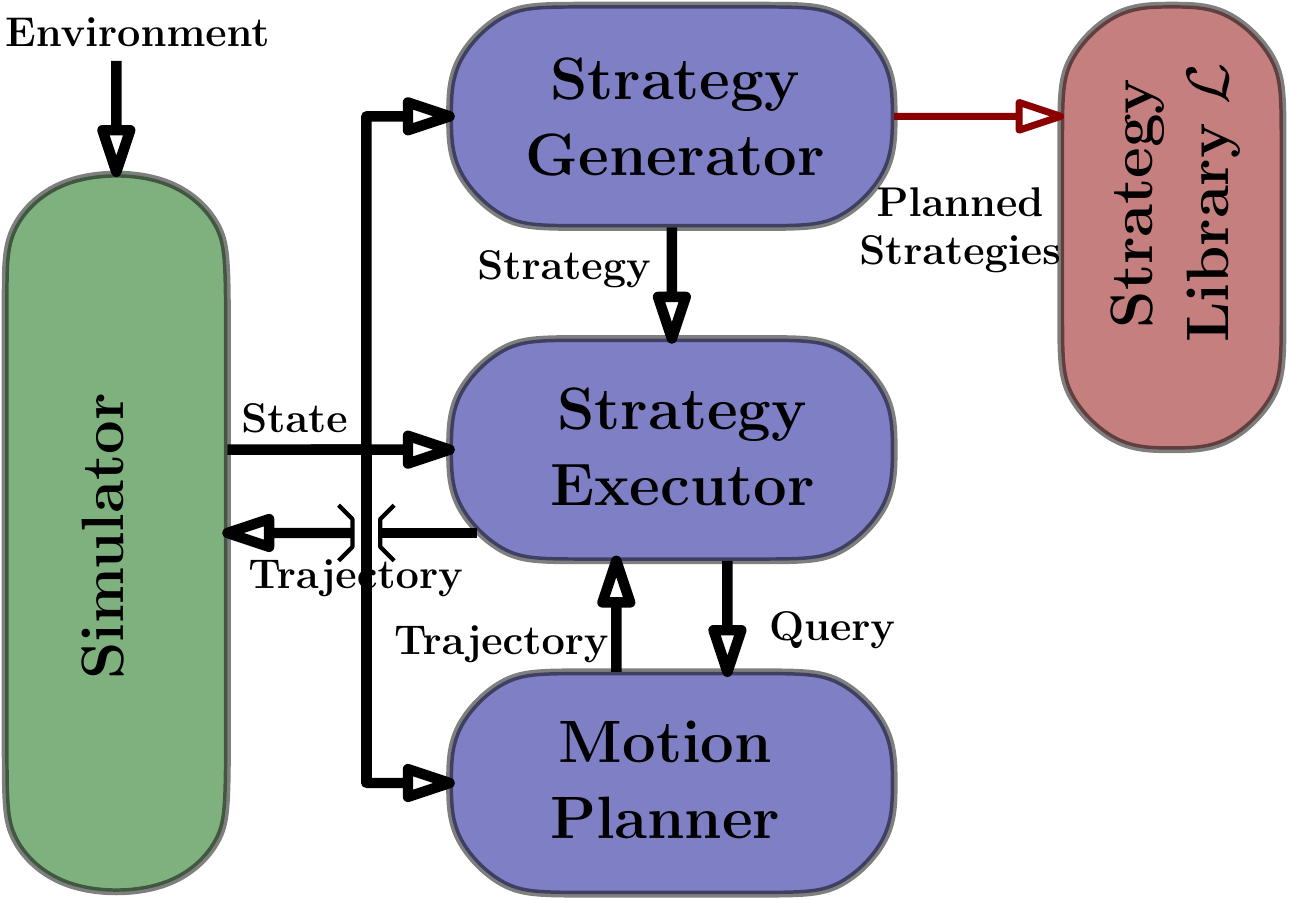}
	\caption{Offline phase, step 1---generation of strategy library.}
	\label{fig:sys1}
  \end{subfigure}
\hspace{1mm}
  \begin{subfigure}{.31\textwidth}
    \centering
    \includegraphics[width=0.8\linewidth]{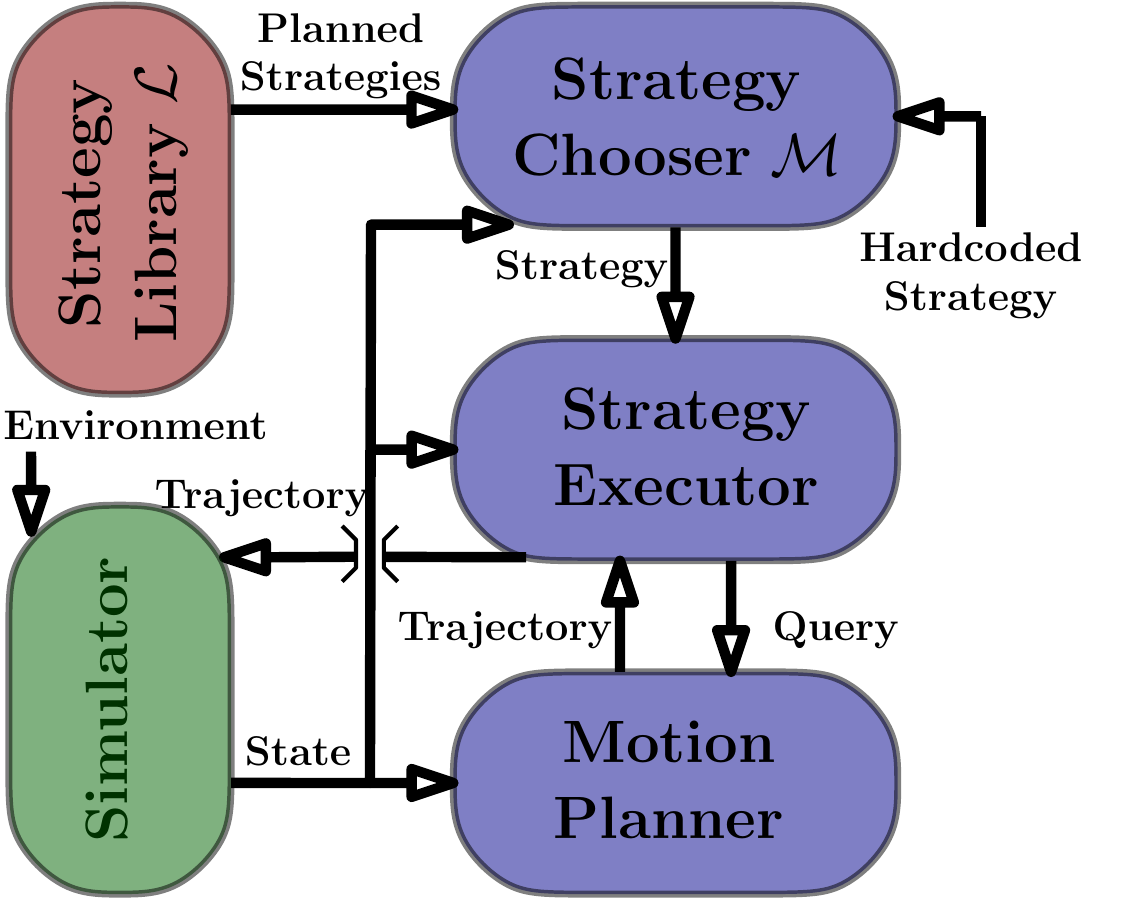}
	\caption{Offline phase, step 2---learning to choose a strategy.}
	\label{fig:sys2}
  \end{subfigure}
  \begin{subfigure}{.31\textwidth}
  	\centering
    \includegraphics[width=0.65\linewidth]{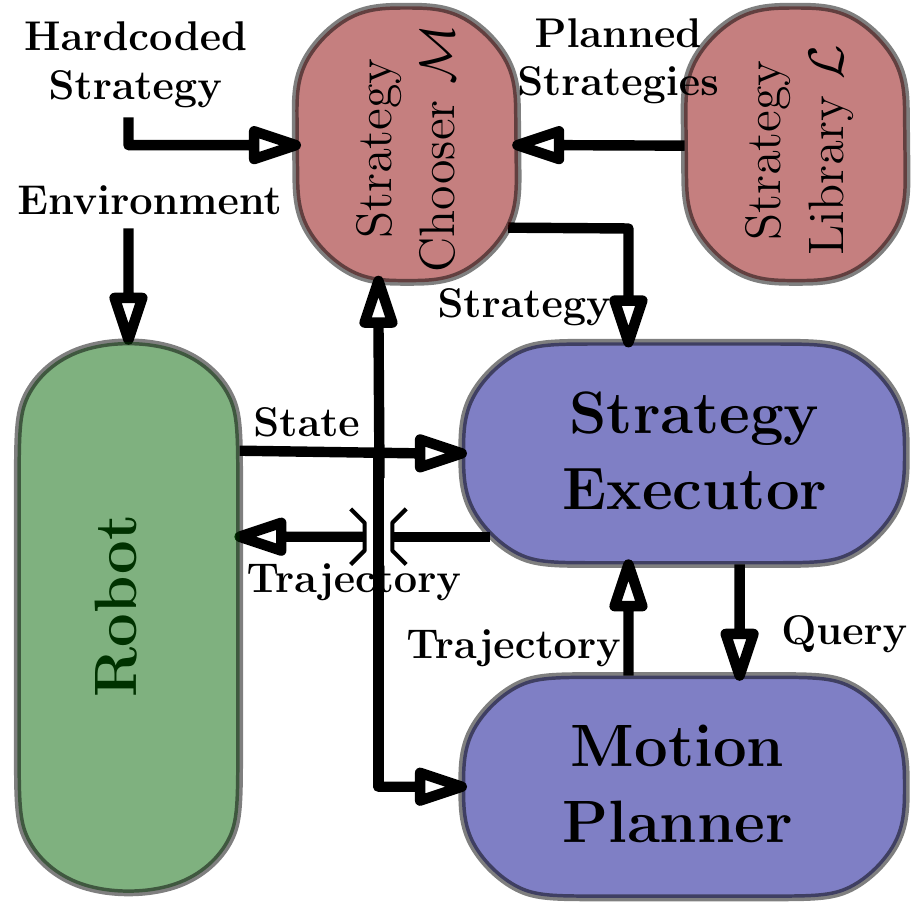}
    \caption{Online phase.}
	\label{fig:sys3}	
  \end{subfigure}

\caption{Truck unloading PLR framework. The offline phase
is done completely in simulation and the online phase can be executed
either on the real robot or in the simulation. Offline phase has two
steps: strategy generation, and learning to choose the most-appropriate
strategy. Online phase involves executing the chosen strategy
given current world state.
PLR has the following components.
(1)~The Motion Planner, used in both phases, generates collision-free trajectories for the robot.
(2)~The Strategy Executor, used in both phases, is tasked with executing the strategy by instantiating the abstract actions. 
%
(3)~The Strategy Generator, used in Offline phase step 1, generates a \emph{library} $\L$ of strategies by evaluating different abstract actions.
(4)~The Strategy Chooser learns a mapping $\M$ offline which is used in the online phase to decide which strategy to use given current world state.}
\label{fig:sys}
\vspace{-2mm}
\end{figure*}

\subsection{PLR Framework Modules}
\label{sec:framework-modules}
The PLR framework uses an offline phase to design strategies which
are then used in the online, truck-unloading phase. The framework is
depicted in Fig.~\ref{fig:sys}.

\subsubsection{Motion Planner}
\label{sec:motion-planner}

The Motion Planner receives as an input a world state $\W$ as well as a motion-planning query and plans a collision-free trajectory. The planner is capable of planning in both
configuration and Cartesian spaces. All the modes that the motion
planner receives queries for are listed in Table.~\ref{tbl:mp}

In our system, minimizing execution time is of utmost importance to unload a large volume of boxes efficiently.
Additionally, short planning times are desired to enable the framework to rapidly evaluate many potential motions, when determining the best way to accomplish an abstract action.
Finally, planning times and generated plans should be \emph{consistent}. Namely, similar queries should roughly take the same amount of planning time and result in similar plans.
Unfortunately, sampling-based planners such as RRT~\cite{LK01} 
often have a large variance in their planning times and solutions due to the stochastic nature of the algorithm. 

To this end, we chose ARA*~\cite{LGT04} as our planner.
ARA* is an anytime heuristic search-based planner which
tunes solution optimality based on available search time.
Specifically, it computes an initial plan quickly and refines its quality as time permits.
Our search space consists of a uniformly-discretized state lattice with motion primitives~\cite{Cohen2010SearchbasedPF}, which are short, kinematically feasible motions for the robot. 
It also employs adaptive motion primitives that rely on an analytical IK solver similarly to the ones described in~\cite{Cohen2010SearchbasedPF} 
To produce efficient-to-execute paths, we use a cost function that approximates the time to perform each motion primitive.
The heuristic function that we use for the ARA* search is the
Euclidean distance to the goal pose for the arm end-effector, and
Euclidean distance in joint space for nose and base.
The planner checks for robot self-collisions using mesh-to-mesh
collision checking and checks for robot collisions with boxes and
trailer using Octree collision geometry. We use Flexible Collision
Library~\cite{pan2012fcl} for collision checking.

\subsubsection{Strategy Executor}
\label{sec:strategy-executor}

\begin{figure*}[t]
\centering
    \begin{subfigure}[b]{\textwidth}
        \centering
        \includegraphics[width=.16\linewidth]{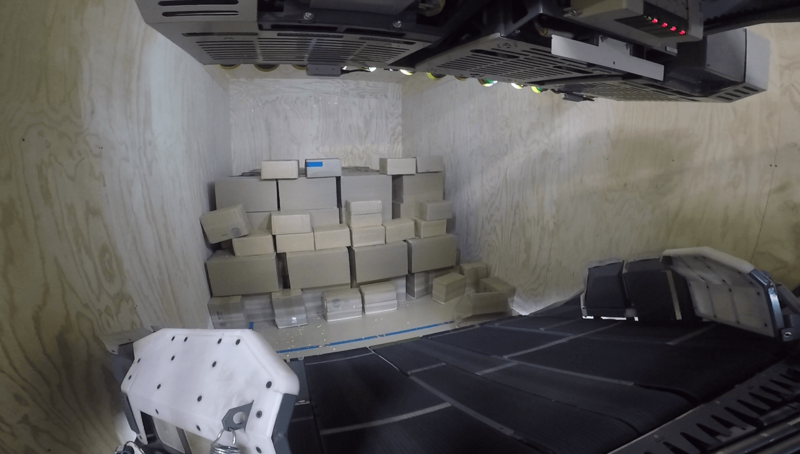}
        \includegraphics[width=.16\linewidth]{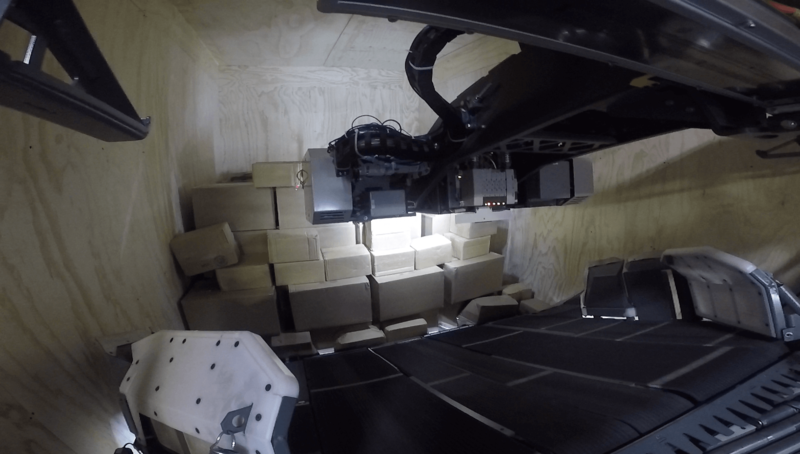}
        \includegraphics[width=.16\linewidth]{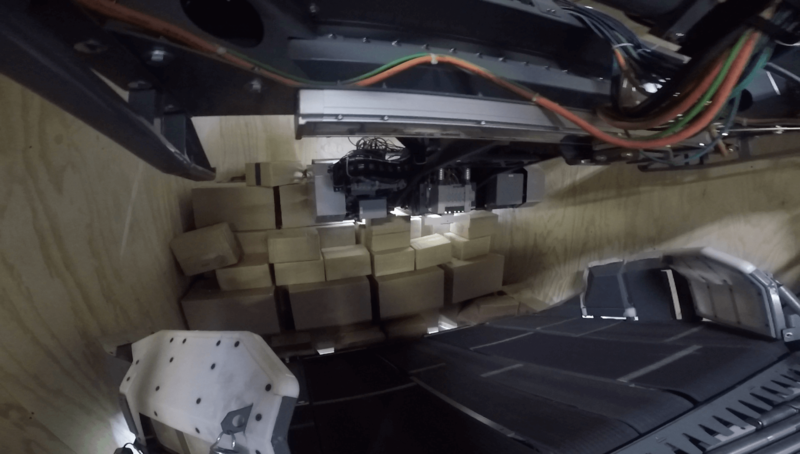}
        \includegraphics[width=.16\linewidth]{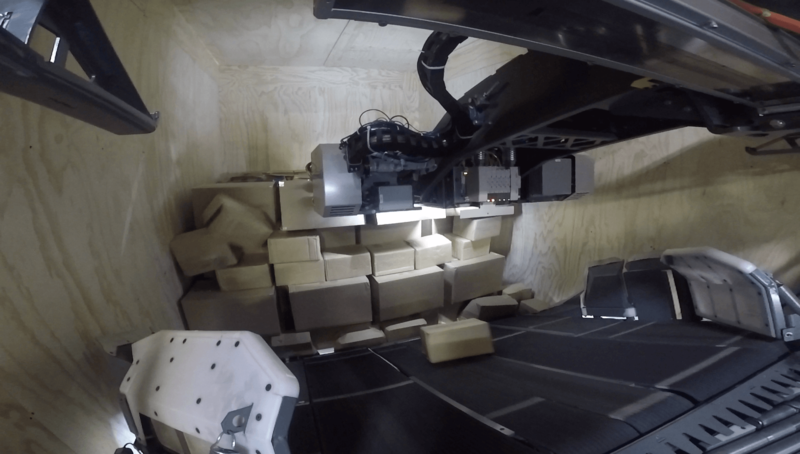}
        \includegraphics[width=.16\linewidth]{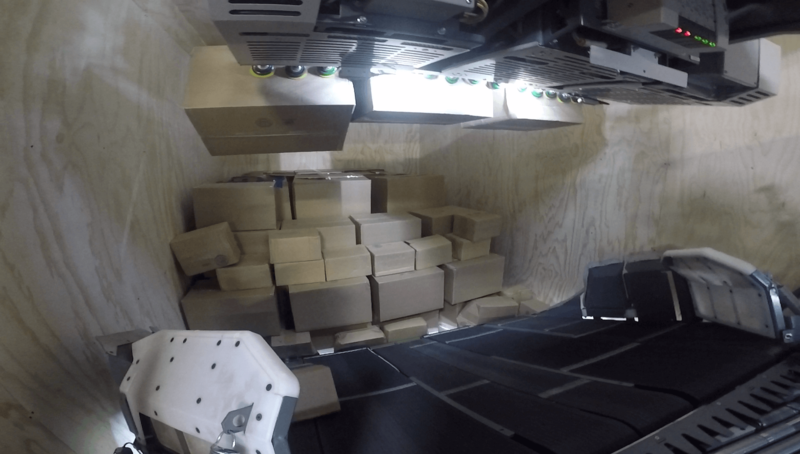}
        \includegraphics[width=.16\linewidth]{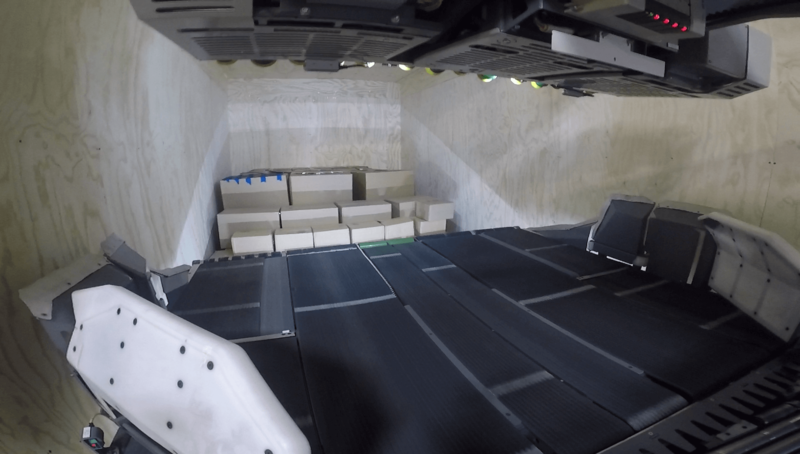}
    \caption{\textbf{The Pick action} is composed of
      six subactions depicted in the pictures sequenced top left to bottom right. Each subaction invokes the suitable
      planning mode. 
      (1)~The nose is lowered down until it touches the
      ground to clear the way for the arm~(M3).
      (2)~The arm is moved to a pre-grasp pose and suction cups are activated~(M2). 
      (3)~The arm is moved to the grasp pose~(M4) 
      (4)~The arm is retracted to a pregrasp pose~(M4). 
      (5)~The arm is moved to the drop-off pose and suction cups are deactivated~(M1).
      (6)~The nose is stowed back to its home configuration (M3).
      In addition to these six subactions the robot base also moves back and forth before and after the Pick action (M5 queries).}
    \label{fig:pick-action}
    \end{subfigure}

    \hspace{1mm}

    \begin{subfigure}[b]{\textwidth}
    \centering
        \includegraphics[width=.16\linewidth]{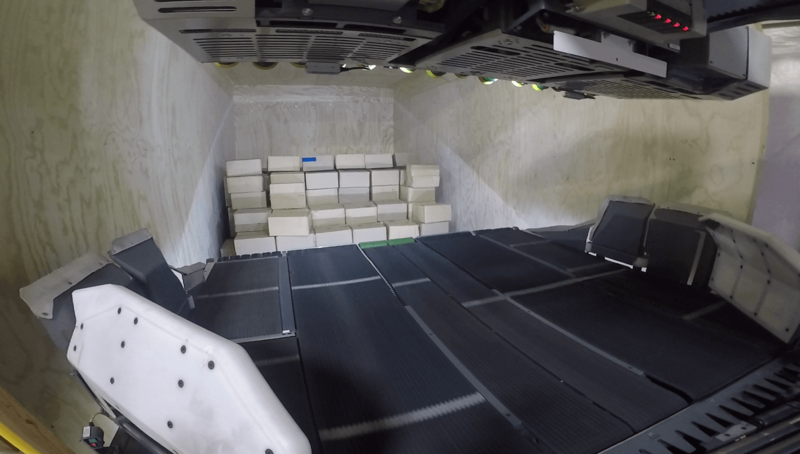}
        \includegraphics[width=.16\linewidth]{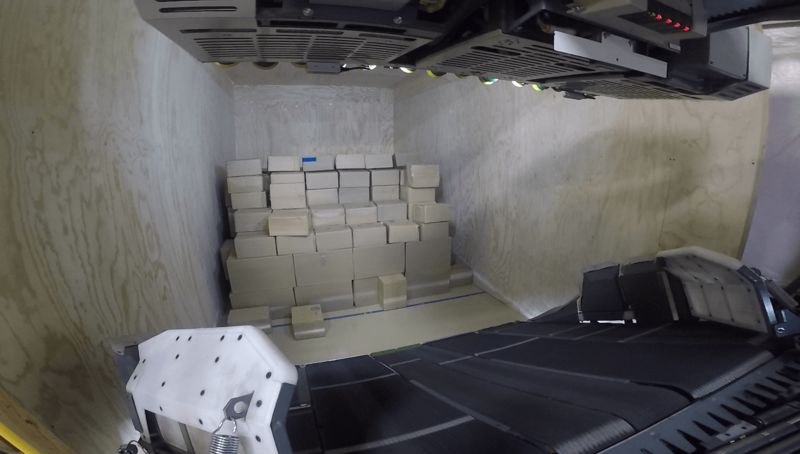}
        \includegraphics[width=.16\linewidth]{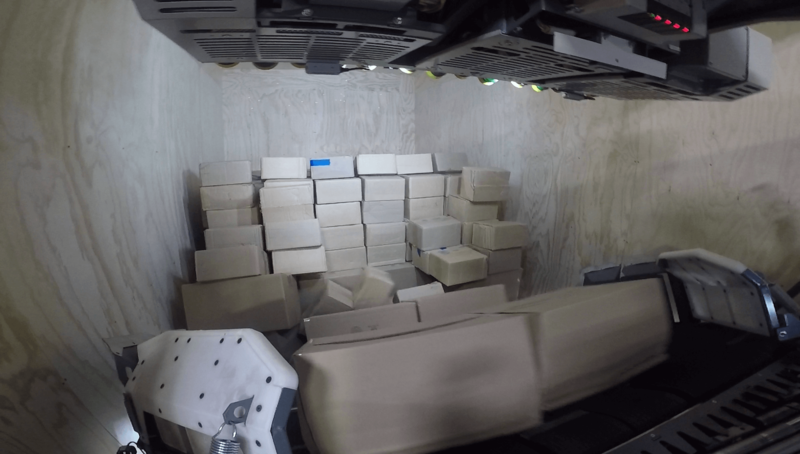}
        \includegraphics[width=.16\linewidth]{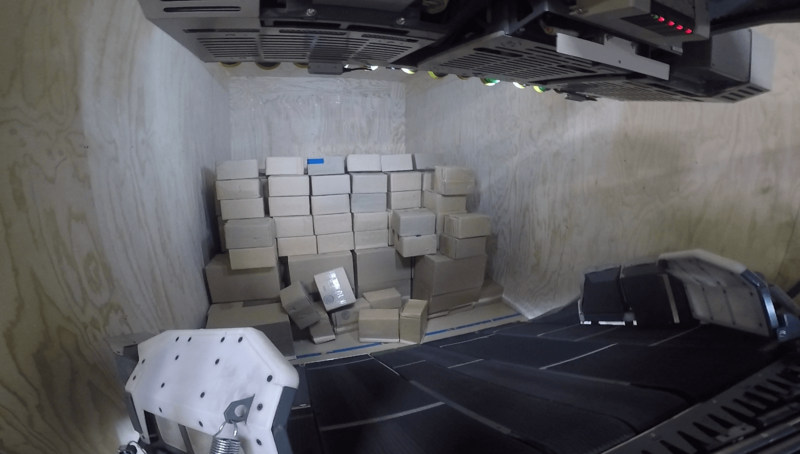}
        \includegraphics[width=.16\linewidth]{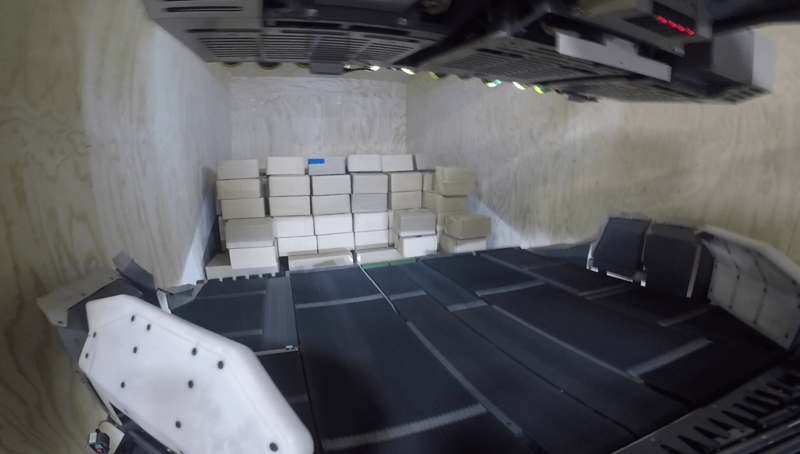}
    \caption{\textbf{The Sweep action} is composed of five subactions sequenced top left to bottom right. 
    (1)~The robot backs up to allow the nose to be lowered~(M5) 
    (2)~The nose is lowered down until it touches the ground~(M3).
    (3)~The base is moved forward to perform the sweep~(M5).
    (4)~The robot backs up to the home configuration~(M5). 
    (5)~The nose is stowed back to the home configuration~(M3).}
    \label{fig:sweep-action}
    \end{subfigure}

\caption{Strategy Executor---Examples of action execution at runtime.}
\label{fig:actions}
\vspace{-5mm}
\end{figure*}

The Strategy Executor executes a strategy by continually evaluating the current state of the world~$\W$ as received from
the robot's sensors and instantiating abstract actions (Fig.~\ref{fig:absact}) into specific motion-planning queries which are then executed by the robot (in simulation or real world).
This is done until the strategy is completed or an alternative one is provided by the Strategy Chooser.
The robot is capable of executing two high level actions---Pick and
Sweep, both depicted in Fig.~\ref{fig:actions}. Pick action works well
in unloading boxes in structured walls, whereas the Sweep action is
geared towards boxes lying in unstructured piles on the trailer
floor. In general, Sweep action has higher mean unloading rate than
Pick action.

Once the Strategy Executor receives a strategy (see Fig.~\ref{fig:treesol} for an example), the execution starts from the root of the strategy tree and goes down, executing actions $u_i$ at each level of the tree, corresponding to the observations $z_i$ made from the world state $\W$.
It instantiates the abstract actions into geometric motion planning queries which are sent to the Motion Planner. It then receives the planned trajectories from the Motion Planner and sends them to the robot controller or the simulator for execution.

\begin{figure}[t]
 \centering

    \includegraphics[width=0.45\textwidth]{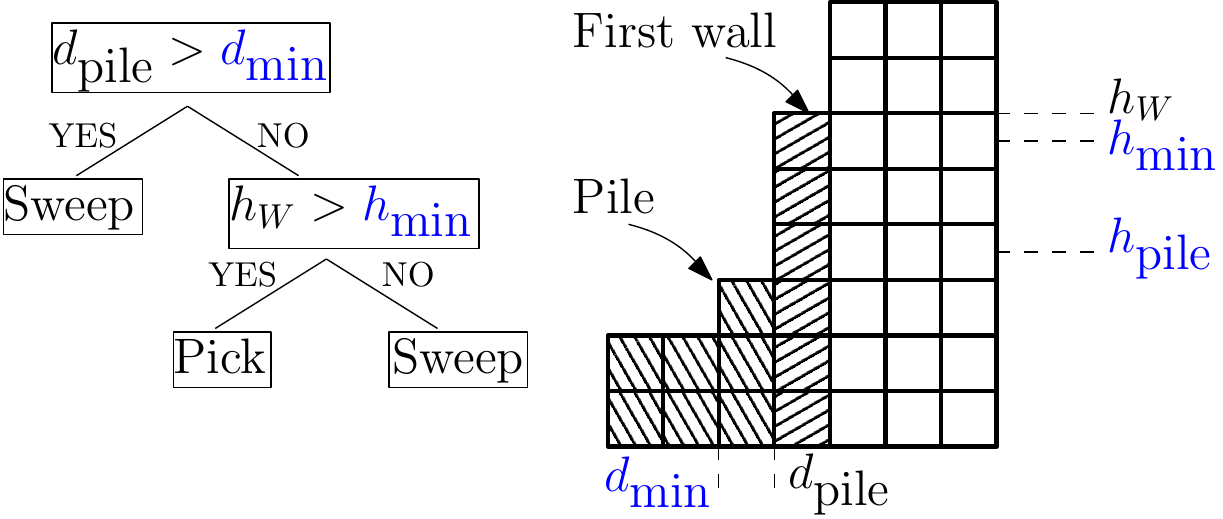}

\caption{The hardcoded strategy for deciding between Pick and Sweep
  actions uses three parameters $h_{\text{pile}}, h_{\text{min}}$ and
  $d_{\text{min}}$ (highlighted in blue). The right figure shows the side view of a possible
  configuration of boxes. Boxes are categorised as either ``pile'' or
  ``wall'' based on the threshold height $h_{\text{pile}}$. The
  decision tree in the left explains the logic. If the depth of the
  pile $d_{\text{pile}}$ is greater than the threshold depth
  $d_{\text{min}}$, it chooses to sweep. Otherwise it checks whether
  the height of the first wall $h_{\text{W}}$ is greater than the
  threshold height~$h_{\text{min}}$ and if so, performs a Pick.
  Pick action is instantiated by choosing a pick point preferring boxes that are closer to the robot and as high as possible. Sweep action is instantiated by choosing the depth of the sweep to be  $d_{\text{pile}}$.}
\label{fig:hardcoded}
\vspace{-7mm}
\end{figure}

\subsubsection{Strategy Generator}
\label{sec:strategy-generator}

The Strategy Generator is tasked with precomputing a set of robust and efficient strategies~$\L$ for a small set of sampled environments such as those depicted in Fig.~\ref{fig:trailers}.
Note that the generator only has access to the perceived world state $\W$,
but not to the ground truth $\W^t$.
Box masses, which affect the environment's dynamics
and the choice of best strategy,
are unknown to the generator. To understand why box masses matter,
consider a truck in which boxes are heavy. In such a case,
picking large number of boxes at once might not be feasible as
suction grippers have limited strength,
so a better strategy might be to pick few boxes at a time.
This robust planning problem under uncertainty, also known as belief
space planning, can be formulated as a POMDP~\cite{KLC98,K15}.
In this work, we employ a recent search-based POMDP solver, Partially Observable Multi-Heuristic Dynamic Programming (POMHDP), which incorporates multiple heuristics from the domain knowledge, particle filtering for belief representation, and iterative forward search for faster convergence \cite{KSL19}.

The goal for this belief space planning problem is to unload all boxes in the truck.
Heuristic functions are used to guide the search to the goal, so that the algorithm does not need to exhaustively evaluate all the possible action sequences.
For this application, we used three heuristic functions, which are
simple yet effective in practice as an ensemble:
(1)~a constant value if there are any boxes left in the truck,
(2)~the total number of boxes left in the truck, and
(3)~the maximum height of the box stacks in the truck.

The action set for the Strategy Generator consists of~12 parameterized abstract actions (Fig.~\ref{fig:absact}).
The observation set consists of 18 semantic observations which describe the world state at a semantic level,
e.g., \texttt{BoxPileLowLeft} is observed if the pile of boxes in front of the robot is lower than a threshold and its topmost box is in the left-hand side.

The belief, i.e., probabilistic estimation of the true state, is
represented by a set of particles, encoding a probability mass function.
Each particle is a sampled world state, thus, the masses of the boxes may differ between particles.
For the given initial belief, POMHDP evaluates the possible sequences of abstract actions and observations (up to a finite horizon), and returns a decision tree with the best abstract actions and the corresponding observations.

\begin{figure}
\centering
\includegraphics[width=0.49\linewidth]{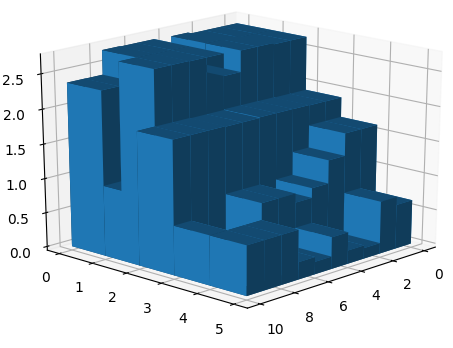}
\includegraphics[width=0.49\linewidth]{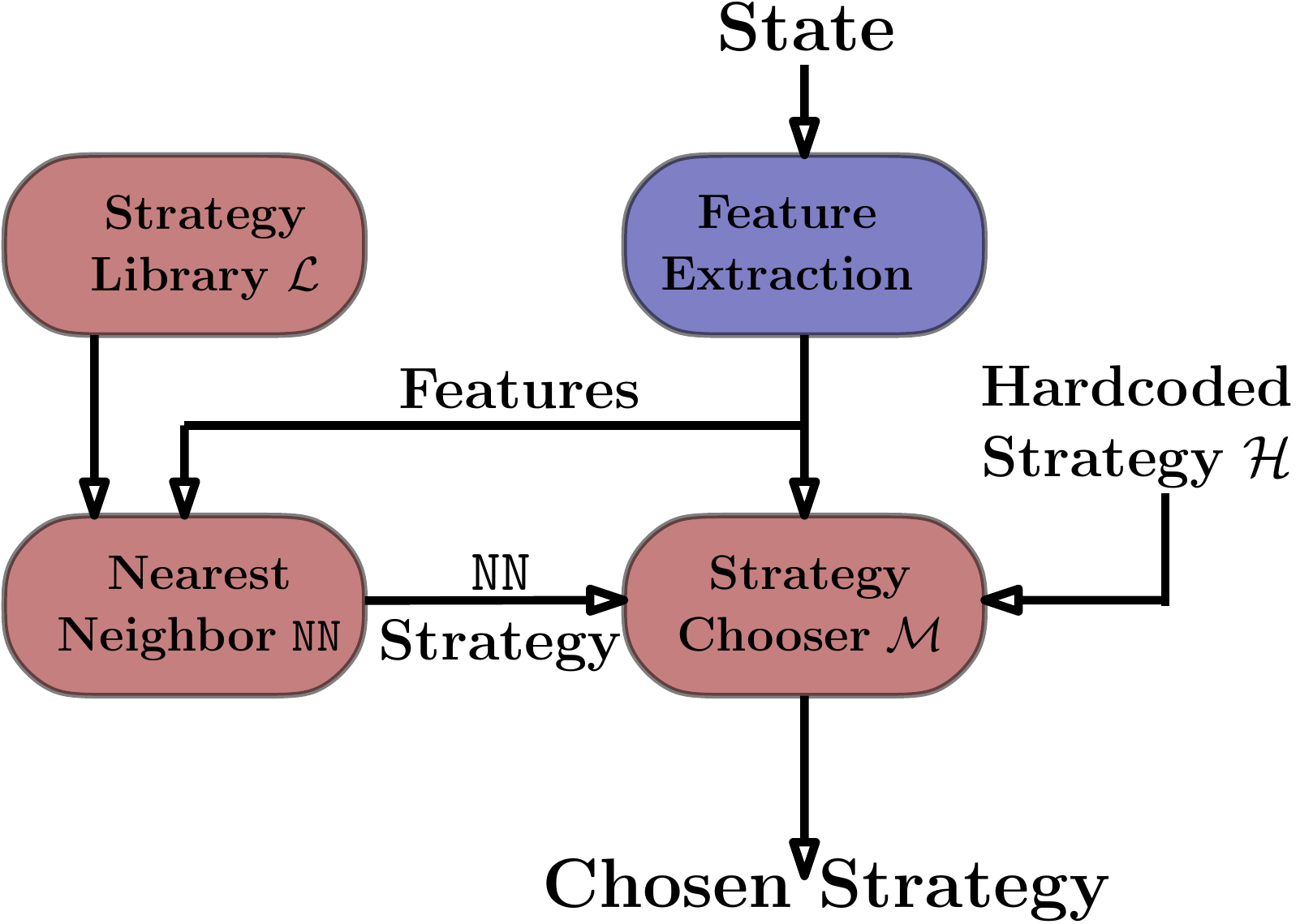}
  \caption{(left) World state discretization to compute features
    (right) Learning framework.
}
  \label{fig:learning-framework}
  \vspace{-8mm}
\end{figure}

\subsubsection{Strategy Chooser}
\label{sec:strategy-chooser}
The Strategy Chooser is
tasked with learning a mapping $\mathcal{M}$ that is used in the online phase to choose the
strategy to execute for a given $\W$. 
As described before, the Strategy
Generator precomputes robust and efficient strategies $\L$ only for a
small set of world states and during the online phase the system encounters new world states and needs to generalize one of the
precomputed strategies in  $\L$ to ensure high throughput.

One solution to
tackle this problem is to use a
\textit{hardcoded strategy} $\H$,
created by a human expert
that encodes domain knowledge. The hardcoded strategy
employed in our experiments is described in Fig.~\ref{fig:hardcoded}.
However, encoding satisfactory behavior by a single strategy, hardcoded or automatically generated, for the large set of environments that will be encountered is extremely challenging. In contrast, we present a learning-based solution to
train~$\M$ as a binary classifier that chooses either the
hardcoded strategy $\H$ or one of the precomputed strategies in
$\L$ which is chosen
using a simple 1-nearest neighbor classifier~$\NN$. After training, $\M$ chooses a pre-computed strategy for a $\W$ that is
similar to sampled states during strategy generation, and chooses
$\H$ for world states that are far. This ensures that our framework
never performs worse than $\H$, and in some cases, significantly
better than~$\H$.

To aid in generalization, we extract features of
the world state using a set of hand-designed features including heights of each column in the wall of boxes, and difference in heights of successive columns where the world state is discretized into a 3D grid with a fixed discretization length. The 3D discretization is shown in Fig.~\ref{fig:learning-framework} (left) for an example world state.
The features are inspired from similar features that were used to learn efficient policies to solve the game of Tetris~\cite{thiery2009improvements}. In our experiments, the number of features extracted is $85$. 
The binary classifier~$\M$, in our
case, is a support vector machine~\cite{cortes1995support} with a Gaussian kernel. The
entire learning framework is
summarized in Fig.~\ref{fig:learning-framework} (right).

Observe that the task of truck unloading is a sequential decision-making
problem where current actions influence future world states. 
To keep the sample complexity required tractable, we formulate it as a supervised learning problem where the objective at every time step is to predict the strategy that will unload the largest number of boxes, given the current world state. In spite of the simplification, we find that this objective results in good long-term unloading performance.
To train~$\M$, we need large amounts of data~$\D$
consisting of pairs of world states and corresponding best strategies. 
This data is obtained in the offline phase iteratively by initializing a random world state, executing (1)~the strategy in~$\L$ as predicted by the nearest-neighbor classifier $\NN$ using the designed features and (2)~the hardcoded strategy~$\H$.
This allows us to determine which strategy
performs the best in terms of box unloading rate for the initialized world state. 
Once the best strategy is recorded in~$\D$, we retrain~$\M$
using~$\D$ and execute the strategy predicted by $\M$ to obtain the subsequent world state. 
Crucial to such a data-collection procedure is the capability to reset the world state in the simulator to execute both strategies. We continue this training for a large number of iterations to obtain~$\M$ that achieves high
performance on a held-out validation dataset. This iterative training
procedure is similar to DAgger~\cite{ross2011reduction}.

\section{Experimental analysis}
\label{sec:eval}

\begin{figure*}[t]
\centering
  \begin{subfigure}{.19\textwidth}
    \centering
    \includegraphics[clip,trim=0cm 0cm 6cm 0cm,width=\linewidth]{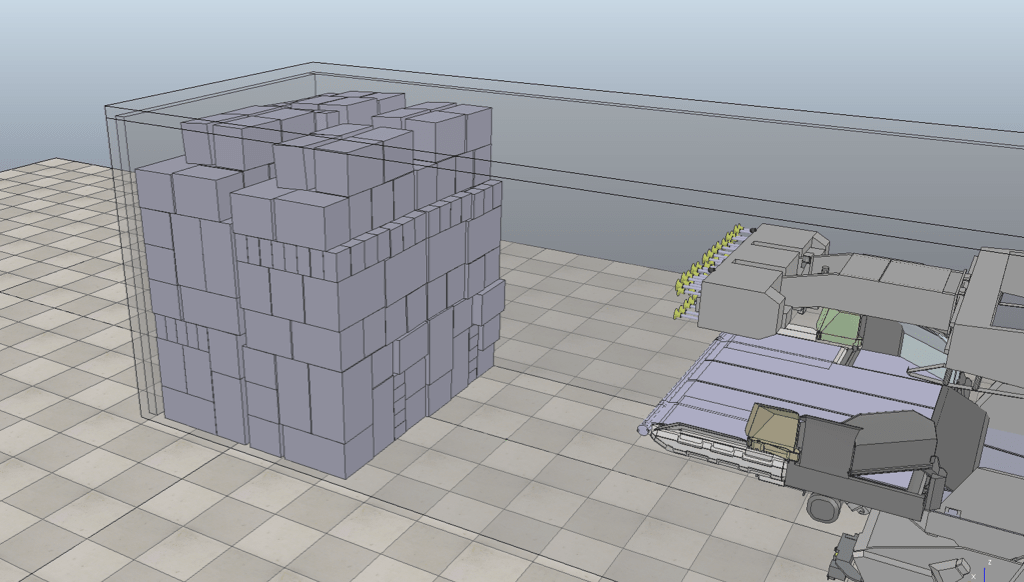}
	\caption{Env.~A1.}
  \end{subfigure}
  \begin{subfigure}{.19\textwidth}
    \centering
    \includegraphics[clip,trim=0cm 0cm 6cm 0cm,width=\linewidth]{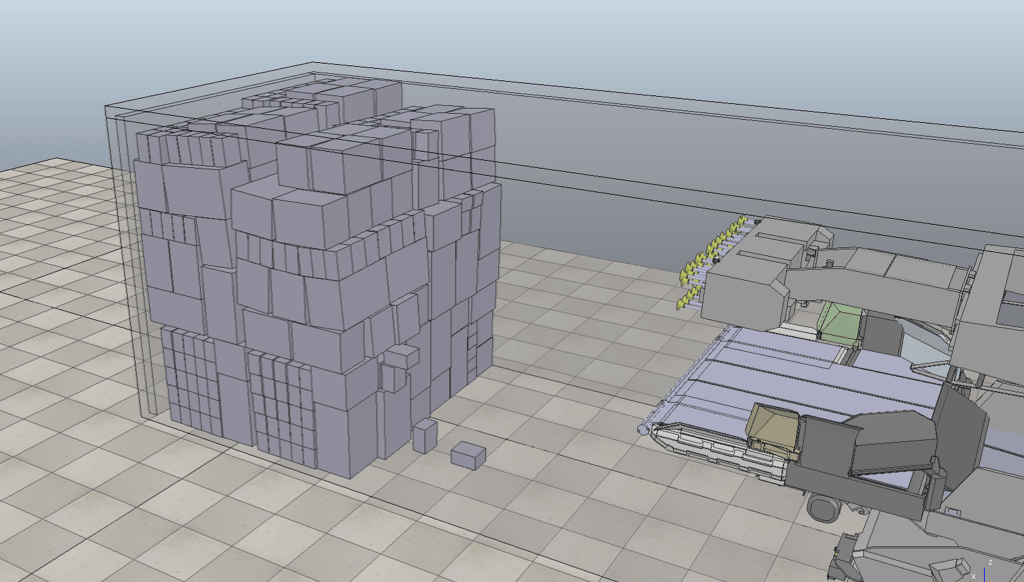}
	\caption{Env.~A2.}
  \end{subfigure}
  \begin{subfigure}{.19\textwidth}
    \centering
    \includegraphics[clip,trim=0cm 0cm 6cm 0cm,width=\linewidth]{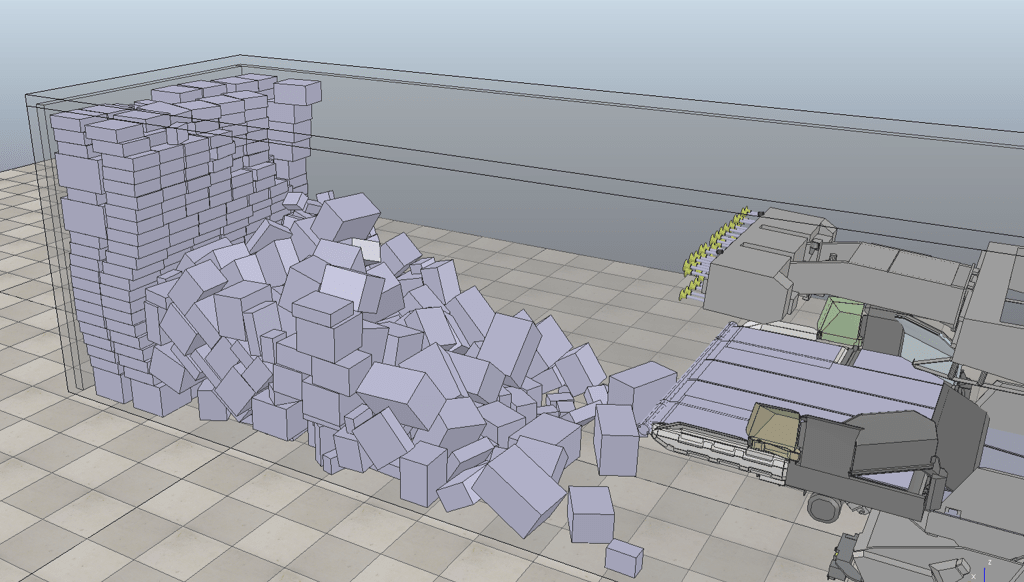}
	\caption{Env.~B1.}
  \end{subfigure}
  \begin{subfigure}{.19\textwidth}
    \centering
    \includegraphics[clip,trim=0cm 0cm 6cm 0cm,width=\linewidth]{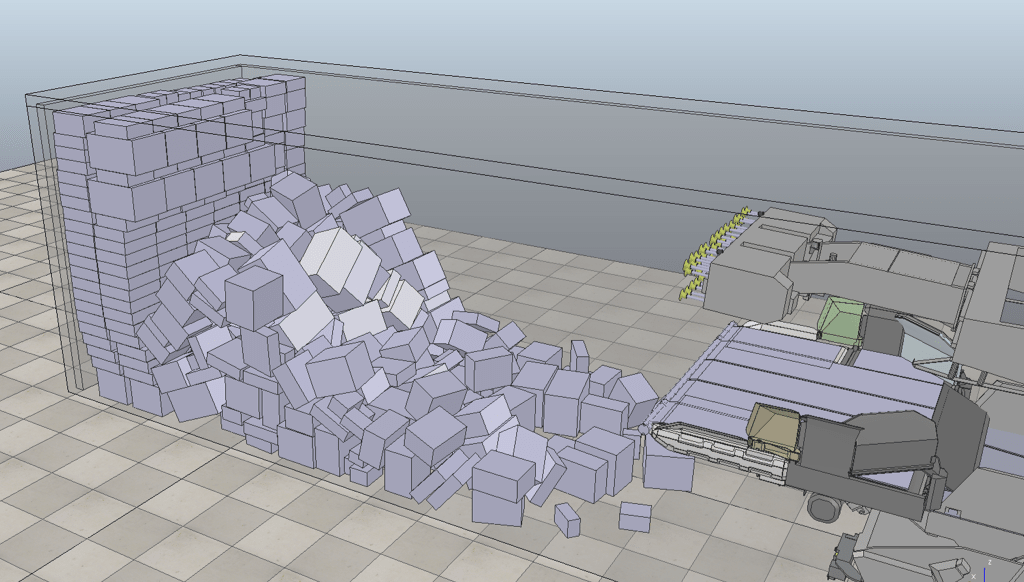}
	\caption{Env.~B2.}
  \end{subfigure}
   \begin{subfigure}{.19\textwidth}
    \centering
    \includegraphics[clip,trim=0.9cm 0cm 0.9cm 0cm,width=\linewidth]{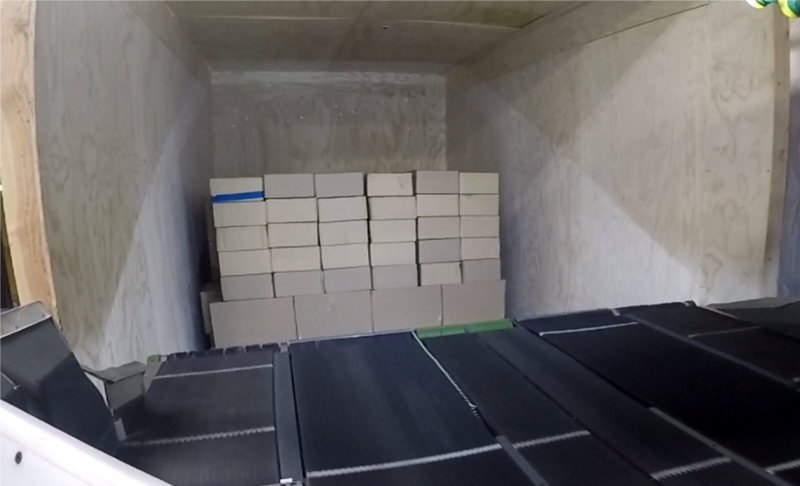}
	\caption{Env. real world}
	\label{fig:realworld}
  \end{subfigure}
\caption{Screenshots of the environments in simulation and real world for experimental analysis.}
\label{fig:scene}
\vspace{-3mm}
\end{figure*}

We present experimental analysis that highlights our design choices.
All results were obtained using a \Cpp implementation of our \plr architecture. We used Intel Xeon Gold 3.40GHz CPU machine for all our experiments. All simulation environments used for evaluation were motivated by real-world
scenarios (see Fig.~\ref{fig:trailers}).
To verify the validity of our basic modules, the Strategy Executor and
Motion planner were also run on the real robot with the 
hardcoded strategy described in Fig.~\ref{fig:hardcoded}.
In our simulated experiments, we used four different environments termed Env.~A1,~A2 and Env.~B1,~B2 (see Fig,~\ref{fig:scene}).
Env.~A1 and~A2 contain nicely-stacked boxes with varying masses and different sizes.
In contrast, Env.~B1 and~B2 contain unstructured piles of boxes on the truck floor with a wide variety of box masses and sizes, which makes the unloading problem significantly harder. 
Env.~A1 and~A2 are significantly different from Env.~B1 and~B2 and thus would require very different strategies for efficient unloading.
We used Env.~A1 and~B1 to generate the library $\L$ and train~$\M$,
according to the procedure described in
Sec.~\ref{sec:strategy-generator}, \ref{sec:strategy-chooser}. Env.~A2 and~B2 were used only to evaluate the system's performance. Video of our physical robot experiments can be found at \url{https://www.youtube.com/watch?v=hRiRhS0kgSg}.

\subsection{Motion planner.}
To demonstrate the extent to which each
motion planning mode is invoked by the Strategy Executor within the
PLR, we start by reporting on the number of queries per mode and the
respective planning times and failure rate for the online phase of the
PLR (Table~\ref{tbl:mp}). The timeout for the motion planner was set to five seconds.
To execute the Pick and Sweep actions the executor makes multiple calls to these modes as detailed in Fig.~\ref{fig:actions}. We also report the overall planning times for the Pick and Sweep actions (Table~\ref{tbl:actions}) which come from accumulated planning times for all the subactions.

As we can see, most planning queries are for relatively low-degree of
freedom planning problems (less than five) and the planner efficiently
finds the plans with almost no timeouts. The only exception is
planning the coordinated motion for the arm and base (M2) which takes
on average almost 1.1 seconds. M5 is the most frequent planning query
as the robot base moves back and forth in each action.

\begin{table}[t]
\begin{center}
\resizebox{\columnwidth}{!}{
\begin{tabular}{|l|cccc|}
\rowcolor[HTML]{C0C0C0} 
\hline 
Plan mode & DOF & No. queries & Planner times [ms] & Failures [\%] \\

M1: Arm            & 4 & 10.9 $\pm$ 5.7 & 90.9 $\pm$ 12.8 & 0 $\pm$ 0\\
M2: Arm + Base     & 5 & 11.1 $\pm$ 5.9 & 1104.8 $\pm$ 708.1 & 0.2 $\pm$ 0.4\\
M3: Nose           & 3 & 47.6 $\pm$ 30.8 & 374.6 $\pm$ 535.1 & 2.2 $\pm$ 1.1\\
M4: Arm         & 1 & 22.6 $\pm$ 12.6 & 6.1 $\pm$ 0.7 & 1.4 $\pm$ 2.7\\
M5: Base        & 1 & 154.7 $\pm$ 56.3 & 2.8 $\pm$ 3.6 & 3.2 $\pm$ 5.2\\
M6: Pre-planning   & 6 & 85.2 $\pm$ 31.7 & 70.9 $\pm$ 465.2 & 5.1 $\pm$ 16.1\\
\hline
\end{tabular}
}
\end{center}
\caption{The different Motion-planner modes, corresponding plannning
  DOF and the statistics for the online phase of Env.~A1 averaged over
  10 different runs and reported with standard deviation.
There are different motion planner modes used by the Strategy
Executor. The modes M1-M3 run ARA* search, M4 and M5 are Cartesian
planners and M6 runs Dijkstra's search. The queries for which the
start configuration is in collision, M6 is run as a pre-planning step to find the closest valid state (to snap onto it). This happens quite often as the robot
makes contacts with the environment during operation. Note that the
planning times are to obtain the first solution returned by ARA*.}
\label{tbl:mp}
\vspace{-2mm}
\end{table}

\begin{table}[t]
\begin{center}
\resizebox{\columnwidth}{!}{
\begin{tabular}{|c|cccc|}
\rowcolor[HTML]{C0C0C0} 
\hline 
Type & No. of execs. & Plan time [s] & Exec. time [s] & Rate [boxes/s]\\

Pick      & 16.8 $\pm$ 7.3 & 2.2 $\pm$ 1.7 & 38.9 $\pm$ 6.3  & 0.2 $\pm$ 0.1 \\
Sweep     & 15.3 $\pm$ 5.5 & 1.7 $\pm$ 3.3 & 59.9 $\pm$ 31.4 & 0.4 $\pm$ 0.5 \\

\hline
\end{tabular}
}
\end{center}
\caption{Strategy executor statistics for Env.~A1 in simulation
averaged over 10 different runs. The total duration of the runs was 26.76 $\pm$ 4.46 [mins]}
\label{tbl:actions}
\vspace{-7mm}
\end{table}

\subsection{Strategy Executor.}
We evaluated the Strategy Executor module in simulation and in the
real world. Table~\ref{tbl:actions} shows the simulation
results for each action, Pick and Sweep, the number of executions,
planning and execution times and the box unloading rate. For Env.~A1,
the frequency of the two actions, Pick and Sweep, is fairly even. The
execution times dominate the planning times for both actions. Note
that even though it takes more time to sweep than to pick, the
mean unloading rate of the Sweep action is about twice as much as the pick
action.

For the real world experiment, we ran the motion planner and the
executor modules with the hardcoded strategy for 10 minutes and 32
seconds. The real world environment is shown in
Fig.~\ref{fig:realworld}. The boxes in the scene vary in sizes. We extract the pile and the walls by fitting planes to the depth values from the point cloud obtained using our sensors. Once we extract the pile and the walls, we use the hardcoded strategy as described in Fig.~\ref{fig:hardcoded}. A
total of 128 boxes were unloaded in the entire run which results in an
unloading rate of 0.2 boxes/sec (see video.)

\begin{figure*}[h]
  \centering
  \includegraphics[width=\linewidth]{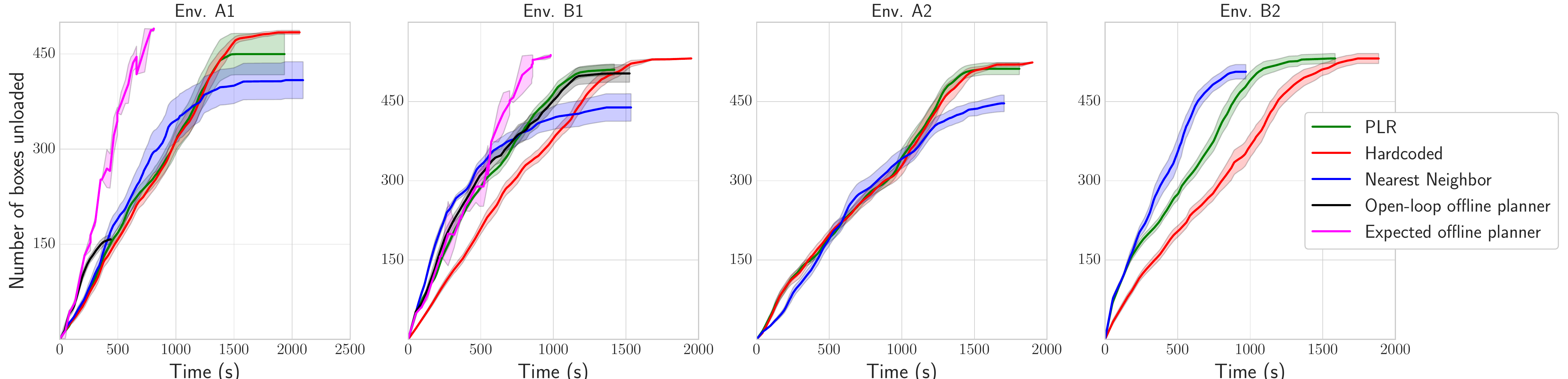}
    \caption{Number of boxes unloaded by different methods on training
     environments A1, B1, and test environments
     A2, B2)
   To evaluate our framework, we compared \plr with several alternatives,
  (1) \textbf{Hardcoded strategy}: Run the hardcoded
  strategy $\H$ (Fig.\ref{fig:hardcoded}) that is
  manually written by a human expert (no offline phase), 
  (2) \textbf{Nearest-Neighbor strategy}:
  Use the strategy predicted by the nearest-neighbor
  classifier $\NN$ from~$\L$ for any environment
  online. Note that here we do not use the binary
  classifier $\M$ or the hardcoded strategy~$\H$, 
  (3) \textbf{Open-loop offline planner:} Run
  precomputed strategies that are generated by Strategy
  Generator. Since a strategy is not a full decision tree, it may
  either reach a point where the strategy has no more actions to
  execute or the strategy encounters a world state that was not
  planned for. This is only possible for training environments, and
  (4) \textbf{Expected offline planner:} Run precomputed
  strategies that are generated by Strategy Generator. However, we reset
  the world state at the end of a strategy to the world state
  corresponding to the start of the next precomputed
  strategy. Although this is infeasible to use in real world, it
  serves as an upper bound on the effectiveness of precomputed
  strategies. This is only possible for training environments.
}
    \label{fig:eval}
\vspace{-5mm}
\end{figure*}
\subsection{Strategy Generator} 
To assess the performance of the Strategy Generator, we report the strategy generation time in the offline phase.
Strategies are generated for Env.~A1 and B1 using five particles, and a planning horizon of six sequential actions.
The approximation factors, $\epsilon_1$ and $\epsilon_2$, in POMHDP are set to 10, respectively.
Note that we employed a high-fidelity slow-speed robot simulator to properly simulate the vacuum suction cups, compliant suction cup joints/plungers, conveyor belts, as well as the complex robot body shapes.
As a result, an abstract action of 40 seconds for execution with the real robot takes about 400-600 seconds for simulation on the aforementioned machines.
To reduce the strategy generation time, we used~25 simulators running
in parallel on separate CPU cores. For Env.~A1 and Env.~B1, it took $3.2 \pm 0.32$ hours and $4.1 \pm 0.87$ hours to generate a strategy, respectively.

\subsection{Overall Framework with the Strategy Chooser}

We start by comparing the performance of the different methods (See
Fig.~\ref{fig:eval} caption) for the case where there is no need for generalization, namely, where the environment used to compute strategies in the offline phases is identical to the one tested in the online phase. In our setup, these are Env.~A1 and~B1.
Fig.~\ref{fig:eval} (left two plots) shows the number of unloaded boxes as a function of time on these environments averaged over~$10$ independent runs (with the shaded region depicting standard error). For both environments, we see that that all the methods other than $\NN$ show comparable performance to the hardcoded strategy~$\H$ which demonstrates that the offline phase effectively computes efficient unloading  strategies. The improvement is more prominent in Env.~B1 than~A1, since the hardcoded strategy is highly tailored for environments such as Env.~A1. Also, note that open-loop execution of precomputed strategies often performs poorly as the unloading process is highly stochastic and we need online adaptation, like the chooser~$\M$, to ensure good performance. The nearest-neighbor strategy~$\NN$ starts with high unloading rates but degrades later as the world states start to diverge from the ones observed in the offline phase due to compounding errors from stochastic dynamics. The chooser~$\M$, however, accounts for this by switching to the hardcoded strategy~$\H$. Thus, it ensures that our system never performs worse than~$\H$, and in some cases, significantly better than~$\H$ (like in Env.~B1 and~B2.)

We observe  similar trends when we evaluate the performance on
environments which have not been seen in the offline phase
(Fig.~\ref{fig:eval} right two plots) Env.~A2 and~B2. 
For Env.~A2, the $\NN$ strategy degrades in the later stage of the unloading task whereas the chooser $\M$ switches to~$\H$. 
In contrast, $\NN$ performs the best in Env.~B2 where the learned strategies generalize well but the chooser $\M$ cannot attain the same performance due to misclassifications. However,~$\M$ still has a significantly better performance than~$\H$, demonstrating the benefit of our \plr framework in terms of generalization.
\vspace{-1mm}






\section{Conclusion and Future Work}
\label{sec:disc-concl}

In this work, we proposed a planning, learning, and reasoning
framework that accounts for uncertainty in the world, generates robust
motion plans offline that are adapted for online real-time execution in
previously unseen environments for automated truck unloading. Our real
world experiments show the real-time performance of our motion
planning and execution modules, while the simulation experiments show
the capabilities of our framework in learning robust offline
strategies that generalize online with better throughput when compared
to hardcoded strategy designed by a human expert.

Future work includes evaluating planned strategies and online adaptation
on the real robot, learning features that are informative in
choosing the appropriate strategy from raw sensor data using
end-to-end learning techniques~\cite{lecun2015deep}, efficient motion
planning for recurring actions such as pick~\cite{ISL19},
and planning with adaptive simulation
accuracy which would enable quicker generation of
strategies~\cite{GSL12, vemula2016path}.


\bibliographystyle{IEEEtran}
\bibliography{bibliography}
\end{document}